\documentclass[runningheads,a4paper]{llncs}

\usepackage{graphicx}{\LARGE }
\usepackage{subfigure}
\usepackage{amsmath,amssymb,amsfonts}{\large }

\usepackage{url}

\newcommand{\keywords}[1]{\par\addvspace\baselineskip
\noindent\keywordname\enspace\ignorespaces#1}

\begin{document}

\mainmatter  

\title{A note on the Variation\\  of Geometric Functionals}

\titlerunning{The Variation of Geometric Functionals}

%
%
\author{Nir Sochen } 
%
\authorrunning{N. Sochen}

\institute{Department of Applied Mathematics,\\
Tel-Aviv University\\
Tel-Aviv, 69978\\
Israel\\
}
%
%

\toctitle{Lecture Notes in Computer Science}
\tocauthor{Authors' Instructions}
\maketitle

\begin{abstract}
Calculus of Variation combined with Differential Geometry as tools of modelling and solving problems in image processing and computer vision 
were introduced in the late 80's and the 90s of the 20th century. The beginning of an extensive work in these directions was marked by works such as 
Geodesic Active Contours (GAC) \cite{GAC}, the Beltrami framework \cite{Beltrami},  level set method of Osher and Sethian \cite{OS}  
the works of Charpiat et al \cite{Charpiat}  and the works by Chan and Vese \cite{CV} to name just a few.  In many cases the optimization 
of these functional are done by the gradient descent method via the calculation of the Euler-Lagrange equations. Straightforward use of the 
resulted EL equations in the gradient descent scheme leads to non-geometric and in some cases non sensical equations. 
It is costumary to modify these EL equations or even the functional itself in order to obtain geometric and/or sensical equations. 
The aim of this note is to point to the correct way to derive the EL and the gradient descent equations such that the resulted gradient descent equation is geometric and makes sense.

\keywords{Calculus of variations, geometric functionals, Geodesic Active Contour, Beltrami framework, Chan-Vese segmentation functional}
\end{abstract}

\section{Introduction}

Differential geometry in image processing and in computer vision started to flowrish in the 90s with works such as those of Charpiat and Faugeras \cite{Charpiat}, 
Paragios and Derich \cite{PD}, the Osher-Sethian level-set method \cite{OS},  the geodesic active contour segmentation algorithm of Caselles et al \cite{GAC}, the Beltrami framework of Sochen, Kimmel and Malladi \cite{Beltrami}, The Chan-Vese segmentation functional \cite{CV} and many more. Many groups further developed and generalized these ideas and techniques to various applications. 

The minimization of these functionals is done by a Gradient Descent (GD) method based on the Euler-Lagrange (EL) equations that result from the variation of the functional. In most of these works the variation of the functional leads to an equation which is NOT geometric and a {\em corrective factor} is added 
in the  GD equation. In other functionals e.g. the Chan-Vese functional,  no factor is added and the result is problematic from a mathematical viewpoint.  

It is the main goal of this note to clarify the problem and to offer a remedy, based on modified inner product,  that overcomes these difficulties. This new way of derivation of the EL and the GD equations solves the appearent problem in the Chan-Vese derivation and leads to a GD equation that makes perfect sense analytically and numerically. .

On one hand the correction that is suggested is not new and is known in principle. It is based on choosing the correct inner product from which one extracts the EL equations. This is known and appear in the past both in the works of Charpiat et al. \cite{Charpiat} and in Yezzi et al. \cite{Yezzi}. On the other hand this known principle was never applied to these popular and much used methods. It is the aim of this paper to bridge between the ``in principle" known mathematics and the ``in practice" wrong applied equations. 

\section{Variation of geometric functionals: The problem}

In order to be specific we present here first the Euler-Lagrange equation and the gradient descent equation that results from the given EL. In later subsections we present  three well known geometric functionals and derive their Euler-Lagrange (EL) equations and their  gradient descent equations and point the problematic equations that result from it.
In general we assume a function $f(p)$, that belongs to $L_2$ say, and a functional that depends on the function and its first derivative(s) $f_p(p)$:
\begin{equation}
S[f]=\int{\cal{L}}(f(p),f_p(p))dp
\end{equation}

The variation is defined as 
\begin{equation}
\delta_\eta S[f]=\lim_{\epsilon\to 0}\frac{S[f+\epsilon\eta] - s[f]}{\epsilon}
\end{equation}
standard computation gives
\begin{equation}
 \delta_\eta S[f] = \int \{\frac{\partial {\cal{L}}(f)}{\partial f}(p) - \partial_p\frac{\partial {\cal{L}}(f)}{\partial f_p}(p)\}\eta(p)dp
\end{equation}
The quantity in the curly  bracket is called the variation of the functional with respect to $f$ and is denoted
$$
\frac{ \delta S[f]}{\delta f} = \frac{\partial {\cal{L}}(f)}{\partial f}(p) - \partial_p\frac{\partial {\cal{L}}(f)}{\partial f_p}(p)
$$

The demand that the function $f$ be a local minimum of the functional $S$  enforce the condition $ \delta_\eta S[f] =0 $ for all possible test functions $\eta$
and yield the Euler-Lagrange equation
\begin{equation}
\frac{\partial {\cal{L}}(f)}{\partial f}(p) - \partial_p\frac{\partial {\cal{L}}(f)}{\partial f_p}(p)=0
 \end{equation}

The gradient descent advance a generic function in a path that minimizes the functional and is guaranteed to arrive to a local minimum that solves the EL equation
\begin{equation}
\frac{\partial f}{\partial t}  =  -  \frac{ \delta S[f]}{\delta f}
 \end{equation}

This completes our short reminder  that serves the purpose of  setting the notations and conventions in an explicit manner such that a closer look in the cases of interest 
can revel the problematics of these equations in the geometric case. 

\subsection{Geodesic Active Contour}
The geodesic active contour (GAC) segmentation functional is a functional over the space of closed curves on the image.  The functional is basically  the length 
of the curve measured with a metric that depends on image features, originally taken as the amplitude of the gradients.  In mathematical terms the curve is described 
parameterically as ${\vec C}(p)=\{X(p), Y(p)\}$  where $p$ is an arbitrary parameter along the curve. One often choose the canonical arc-length parameterization of the curve
${\vec C}(s)=\{X(s), Y(s)\}$ in which the tangent vector $|{\vec T}(s)|=|{\vec C}_s(s)|=1$.   The functional now reads
\begin{equation}
S[{\vec C}] = \int G({\vec C}(s))ds=\int G({\vec C}(p))|{\vec C}_p(p)|dp
\end{equation}
which can be written explicitly as 
\begin{equation}
S[X,Y] = \int G(X(p),Y(p))\sqrt{X_p^2 + Y_p^2}dp
\end{equation}

Now we can apply the calculus of variations to find the EL equations
\begin{eqnarray}
 \frac{ \delta S[X, Y]}{\delta X} &=&  G_X(X,Y)\sqrt{X_p^2 + Y_p^2} - \partial_p\left(G(X,Y)\frac{X_p}{ \sqrt{X_p^2 + Y_p^2}}\right) \cr
  \frac{ \delta S[X, Y]}{\delta Y} &=&  G_Y(X,Y)\sqrt{X_p^2 + Y_p^2} - \partial_p\left(G(X,Y)\frac{Y_p}{ \sqrt{X_p^2 + Y_p^2}}\right) 
  \end{eqnarray}
    and in vectorial notation
 \begin{equation}
 \frac{ \delta S[{\vec C}]}{\delta {\vec C}} =  \left|{\vec C}_p\right| \nabla G({\vec C}) - \partial_p\left(G({\vec C})\frac{{\vec C}_p}{ |{\vec C}_p|}\right) 
  \end{equation}

 If we try now to write down the gradient descent equation, following the general prescription that we detailed above we end with the 
 wrong equation
  \begin{eqnarray}
 \frac{ \partial {\vec C}(p,t)}{\partial t} &=& - |{\vec C}_p| \nabla G({\vec C}) + \partial_p\left(G({\vec C})\frac{{\vec C}_p}{ |{\vec C}_p|}\right) 
  \end{eqnarray}
 This equation is NOT geometric and clearly depends on the parameterization. It is not the desired equation for a model whose functional that was carefully constructed to be independent of the parameterization. Two remedies were suggested for this situation. The first remedy is to work in one preferred parameterization, the arclength.
 This leads to the geometric gradient descent equation 
  \begin{eqnarray}
 \frac{ \partial C(s,t)}{\partial t} &=& - \nabla G(X,Y) + \partial_s\left(G(X,Y)C_s\right) 
  \end{eqnarray}
but then one may ask how come that a reparameterization invariant functional must be treated in  only one specific parameterization? 

The second remedy modifies the gradient descent such that the end result is indeed reparameterization invariant. It is defined as such
\\

\begin{eqnarray}
\frac{ \partial C(s,t)}{\partial t} &=& -   \frac{1}{|C_p|}\frac{\delta S}{\delta C} = 
 \frac{1}{|C_p|}\left( |C_p|\nabla G(X,Y) - \partial_p\left(G(X,Y)\frac{C_p}{ |C_p|}\right) \right)\cr
 &=&  - \nabla G(X,Y) + \partial_s\left(G(X,Y)C_s\right) 
 \end{eqnarray}
 This equation is geometric as desired but the factor $1/|C_p|$ is justified a-posteriori by the resulting equation and not a-priory from basic principles.
 
 No other good explanation is given in the literature as far as we know.

\subsection{The Beltrami flow}
 The Beltrami flow is based on treating images as Riemannian manifolds embedded in higher dimensional Riemannian manifolds.
 For the sake of this paper it is enough to deal with the case that the embedding manifold is flat and is described locally with Cartesian
 coordinate system $X^1,\ldots , X^n$ where $X^i=X^i(x^1, \ldots, x^d)$. The basic functional here is the Polyakov action
 \begin{eqnarray}
 S[X^1,\ldots , X^n, G] = \int dx^d \sqrt{g} \sum_i(\nabla X^i)^TG^{-1}\nabla X^i 
  \end{eqnarray}
  where $g=\det G$.
The EL equation is
\begin{eqnarray}
 \nabla\left(\sqrt{g}G^{-1}\nabla X^i\right)=\sqrt{g} H{\vec N}_i =0
   \end{eqnarray}

   where $H$ is the mean curvature and ${\vec N}$ is the normal to the surface. Clearly this equation is not geometric. One has to divide this equation by ${\sqrt{g}}$ in order to 
   make it a geometric equation.  The same justification is applied in the literature:  The need to have a geometric equation and the legitimacy of the multiplication by a factor. 
   The gradient descent equation is

   \begin{eqnarray}
 \frac{\partial X^i}{\partial t} = -\frac{1}{\sqrt{g}}\frac{\delta S}{\delta X^i} = \frac{1}{\sqrt{g}}\nabla\left(\sqrt{g}G^{-1}\nabla X^i\right)= H{\vec N}_i 
   \end{eqnarray}

\subsection{The Chan-Vese  model}
In this functional that aims to segment an object via a level-set formulation.  The functional, in its original and simple formulation,  is based on a difference between the mean 
grey level inside and outside the object to be segmented.  The functional also penalizes too lengthy boundary of an object. It reads
  \begin{eqnarray}
  \label{CV_functional}
  S[\Phi ; c_1, c_2] &=& \int_\Omega \left[(I(x,y)-c_1)^2H(\Phi(x,y))  +   (I(x,y)-c_2)^2H(-\Phi(x,y))\right]   dxdy \cr 
  &+& \mu \int_\Omega\left |\nabla H(\Phi(x,y)\right | dxdy
   \end{eqnarray}
where $H(s)$ is the heaviside function and $\left |\nabla H(\Phi(x,y)\right | =\delta(\Phi)\left |\nabla\Phi(x,y)\right | $. Straightforward use of calculus of variations yields 
the EL equation 
 \begin{eqnarray}
  \delta(\Phi)\left [(I(x,y)-c_1)^2  -   (I(x,y)-c_2)^2  - \text{Div}\left(\frac{\nabla\Phi}{|\nabla\Phi|}\right) \right] = 0 \cr 
   \end{eqnarray}
   and the gradient descent equation is 
   \begin{eqnarray}
  \frac{d\Phi}{dt} =  \delta(\Phi)\left [-(I(x,y)-c_1)^2  +   (I(x,y)-c_2)^2  + \text{Div}\left(\frac{\nabla\Phi}{|\nabla\Phi|}\right) \right]  \cr 
   \end{eqnarray}
This equation actually means that the zero level-set evolves while all other level-sets are fixed. This is of course impossible without $\Phi$  ceasing being a function which makes no sense.  The way this usually dealt with is by numerical methods that alter this equation such that more level sets are changed at the same time. 
This note aims to remedy this problem of nonsensical gradient descent equation out of completely geometric and sensical functional. 

\section{How to vary a geometric functional}
In this section we will present the process that remedies all the problems that were encountered above. 
The  point is very simple: A non-geometric Euler-Lagrange (or gradient descent) equation can arise from a geometric functional only if somewhere on the way we did a non-geometric 
consideration. Looking back on the process of the derivation of the Euler-Lagrange equation it is clear what is the crucial step: moving from the integeral form of the variation to the 
differential one. The point here is actually well known. The calculus of variation includes not only the functional to be minimized but also the functional space from which 
the minimizer is to be found and the inner product in that (usually Hilbert) space. This is easily demonstrated by examples. 

\subsection{Geodesic Active Contour}

Look at the GAC geometric functional 
\begin{equation}
S[X,Y] = \int G(X(p),Y(p))\sqrt{X_p^2 + Y_p^2}dp
\end{equation}
This is actually a functional on a curve.  The functional is invariant under reparameterization as it should. Looking on the variation we find 
 \begin{equation}
 \label{int-var-GAC}
\delta S[{\vec C}] =\int\left( \left|{\vec C}_p\right| \nabla G({\vec C}) - \partial_p\left(G({\vec C})\frac{{\vec C}_p}{ |{\vec C}_p|}\right) \right)\delta {\vec C}(p)dp
  \end{equation}
The demand $\delta S[{\vec C}]=0$ for any $\delta{\vec C}$ we choose yield the Euler-Lagrange equations. 

The point is that we are treating here the space of functions on the curve with an Euclidean inner product with respect to the parameterisation!  This is of course the source 
of the problem. In fact it is obvious that the inner product for functions on a curve should be
 \begin{equation}
\langle f,\ g\rangle_{\vec C} =\int_{\vec C} f(s)g(s)ds = \int_{\vec C} f(p)g(p)\underbrace{|{\vec C}_p|dp}_{ds}
  \end{equation}
Writing now the variation (\ref{int-var-GAC}) in terms of the geometric inner product we find
\begin{equation}
\delta S[{\vec C}] =\int\underbrace{ \frac{1}{|{\vec C}_p|}\ \left( \left|{\vec C}_p\right| \nabla G({\vec C}) - \partial_p\left(G({\vec C})
\frac{{\vec C}_p}{ |{\vec C}_p|}\right) \right)}_{f(p)}\underbrace{\delta {\vec C}(p)}_{g(p)}\underbrace{|{\vec C}_p|dp}_{ds}
  \end{equation}
  The EL equation now reads
\begin{equation}
f(p)=\nabla G({\vec C}) - \frac{1}{|{\vec C}_p|}\partial_p\left(G({\vec C})\frac{{\vec C}_p}{ |{\vec C}_p|}\right) = \nabla G({\vec C}) - \partial_s\left(G{\vec C}_s\right)=0
 \end{equation}
This equation is indeed geometric as it should! 

\subsection{The Beltrami flow}

Here we reserve the EL equation for the Polyakov action for the embedding of a two-dimensional manifold in higher dimensional space. We take here, for simplicity, 
an embedding space which is flat and is described locally with Cartesian coordinate system.  The space that we consider is the space of functions on the two-dimensional 
manifold and the inner product for two such functions is
\begin{equation}
\langle f,\ h\rangle_{\vec S} =\int_{\vec S} f(\sigma^1,\sigma^2)h(\sigma^1,\sigma^2)\underbrace{\sqrt{g}d\sigma^1d\sigma^2 }_{dA}
  \end{equation}
where $g=\det G$ with $G$ the metric of the manifold. The local parameterization $(\sigma^1,\sigma^2)$ is defined on a local neighborhood. 

Using this inner product we now find
\begin{eqnarray}
\delta S[{X^i]}&=&\int_{\vec S}\underbrace{\nabla\left(\sqrt{g}G^{-1}\nabla X^i\right)}_{\tilde{f}({\vec S})}\underbrace{\delta X^i}_{h({\vec S})}d\sigma^1d\sigma^2\cr
&=& \int_{\vec S}\underbrace{\frac{1}{\sqrt g}\nabla\left(\sqrt{g}G^{-1}\nabla X^i\right)}_{f({\vec S})}\underbrace{\delta X^i}_{h({\vec S})}\underbrace{\sqrt{g}d\sigma^1d\sigma^2 }_{dA}
   \end{eqnarray}
   from which the geometric EL and GD equations follow immediately without any need for correction. 
   
\subsection{The Chan-Vese  model}
This subsection presents the main novelty of this note. We use the same idea of applying a geometric  inner product  in the process of the calculus of variation of geometric 
functionals. In the previous subjections we re-derived known flows and explained the appearance of  factors that were added previously without explanation  while in this subsection 
we show how this reasoning changes substantially the EL and gradient descent  equations of the Chan-Vese functional such that the resulted GD is completely sensible from a mathematical and numerical viewpoints.   

We start by noting that the Chan-Vese functional is a functional of curves in 2D.  It is written via a level-set function $\Phi(x,y)$ on the 2D plain. In terms of the level set function we find
\begin{eqnarray}
\langle f, g\rangle_C = \int_C f(s)g(s)ds = \int_\Omega \tilde{f}(x,y)\tilde{g}(x,y)\underbrace{|\nabla H(\Phi(x,y))|dxdy}_{ds}
   \end{eqnarray}
where the tilde functions are 2D extensions of the original functions on the curve.  Note also that $|\nabla H(\Phi(x,y))| = \delta(\Phi)|\nabla\Phi| $ . With the help of these identities we find that 
the variation of the functional eq. (\ref{CV_functional}) is 
 \begin{eqnarray}
  \label{CV_variation}
  \delta S= \int_\Omega \delta(\Phi)\left\{ (I-c_1)^2  -  (I-c_2)^2   -  \mu\rm{Div}\left( \frac{\nabla\Phi}{|\nabla\Phi |}\right) \right\} \eta(x,y)  dxdy 
   \end{eqnarray}
   where $\eta$ is the variation of $\Phi$.   Rerranging according to the geometric inner product we finally find
\begin{eqnarray}
  \label{CV_variation}
  \delta S= \int_\Omega  \underbrace{\frac{1}{|\nabla\Phi| }\left\{ (I-c_1)^2  -  (I-c_2)^2   -  \mu\rm{Div}\left( \frac{\nabla\Phi}{|\nabla\Phi |}\right) \right\}}_{\tilde f} 
  \underbrace{\eta}_{\tilde g} \underbrace{\delta(\Phi)|\nabla\Phi|  dxdy}_{ds} \cr
   \end{eqnarray}
   It is clear now that the delta function $\delta(\Phi)$ belongs to the measure of integration on the curve i.e. $ds$ and not to the EL equation.  Note also that the EL equation has
   now a factor $1/|\nabla \Phi |$ which comes from the measure $ds$ as well.  The correct gradient descent equation now reads
    \begin{eqnarray}
  \frac{d\Phi}{dt} =  \frac{1}{|\nabla\Phi| }\left [-(I(x,y)-c_1)^2  +   (I(x,y)-c_2)^2  + \mu\text{Div}\left(\frac{\nabla\Phi}{|\nabla\Phi|}\right) \right]  \cr 
   \end{eqnarray}
   which moves ALL the level lines of $\Phi$ and thus makes perfect sense. 
   
   \section{summary}
We showed in this note how to use the calculus of variation for geometric functionals. In particular we showed that a care should be taken regarding the inner product that is used to extract 
the Euler-Lagrange equations. The fact that the choice of inner product in the relevant Hilbert space has an impact on the EL and GD equations is not new \cite{Yezzi,Charpiat}. Yet, it was used in the past to accelerate convergence of the GD \cite{Yezzi} and for the analysis of shapes \cite{Charpiat}. In this paper we apply it for geometric functional that are used in image processing and it is shown that in order to get an appropriate geometric EL and GD equations one has to use a geometric inner product for the functions which are defined over the manifold.
Such an inner product should take into account the length, area or volume elements of the integration. It is shown that using the right inner product we can recover known geometric 
flows  such as the geodesic active contours and the Beltrami flow with no hand waiving a-posteriori multiplication of extra factors.  This assesses the correctness of this approach. It is then applied 
to the Chan-Vese segmentation functional with a modified EL and gradient descent equations. It is shown that the delta function that cause many conceptual and numerical problems actually
belongs to the measure of the integral and not to the EL equation. Careful treatment of the inner product for this functional leads to a new gradient equation that is correct conceptually and much
easier on the numerical side.

\end{document}